%% file: main.tex
\documentclass[10pt,twocolumn,letterpaper]{article}

\usepackage[pagenumbers]{wacv} 

\input{preamble}

\definecolor{wacvblue}{rgb}{0.21,0.49,0.74}
\usepackage[pagebackref,breaklinks,colorlinks,allcolors=wacvblue]{hyperref}

\def\wacvPaperID{1794} 
\def\confName{WACV}
\def\confYear{2027}

\title{Probing Geospatial SSL Representations with Environmental Signals}

\author{Rohita Mocharla$^{1,2}$\\
$^{1}$Johns Hopkins Applied Physics Laboratory\\
Laurel, MD\\
{\tt\small nmochan1@jhu.edu}
\and
Vishal M. Patel$^{2}$\\
$^{2}$Johns Hopkins University\\
Baltimore, MD\\
{\tt\small vpatel36@jhu.edu}
}

\begin{document}
\maketitle
\input{sec/0_abstract}
\input{sec/1_intro}

\input{sec/2_related}
\input{sec/3_evaluation}

\input{sec/4_experiment}
\input{sec/5_results}
\input{sec/6_conclusion}
{
    \small
    \bibliographystyle{ieeenat_fullname}
    \bibliography{main}
}
\clearpage
\input{sec/appendix}

\end{document}

%% file: preamble.tex
\usepackage{multirow}
\usepackage{booktabs}
\usepackage{placeins}
\usepackage[T1]{fontenc}


%% file: sec/0_abstract.tex
\begin{abstract}

Self-supervised learning (SSL) is designed to learn generic, transferable representations rather than representations optimized for a single task. Most geospatial benchmarks evaluate representations solely through downstream tasks, providing limited insight into the information encoded within the representation itself. We ask a different question: do SSL representations of satellite imagery preserve statistical associations with environmental variables that co-vary with the imaging process? To answer this question, we probe SSL representations using co-located ERA5 reanalysis variables~\cite{hersbach2020era5}, a global dataset of physically consistent environmental variables, including temperature, precipitation, surface solar radiation, surface pressure, and volumetric soil water. These variables are physically related to the spectral reflectance and radar backscatter recorded by Sentinel-1 and Sentinel-2, making them meaningful evaluation targets despite not being used during SSL pretraining. We complement this probing analysis with intrinsic representation metrics to characterize representation geometry and investigate how these properties relate to downstream performance and the encoding of environmental signals. Using DINO \cite{dino}, MAE \cite{mae}, and MoCo \cite{moco} models trained under identical conditions, we show that representation-level metrics distinguish models with similar downstream benchmark performance, providing complementary information beyond task-driven benchmarks. We further find that the linear accessibility of environmental signals is associated with performance on environmentally dependent tasks in the PANGAEA benchmark \cite{marsocci2025pangaeaglobalinclusivebenchmark}. Finally, we release ERA5~\cite{hersbach2020era5} annotations co-located with the SSL4EO \cite{wang2022ssl4eo} dataset to enable physically grounded representation evaluation for future geospatial foundation models.

\end{abstract}

%% file: sec/1_intro.tex
\section{Introduction}
\label{sec:intro}
\begin{figure*}[!t]
\centering
    \includegraphics[width=0.8\textwidth]{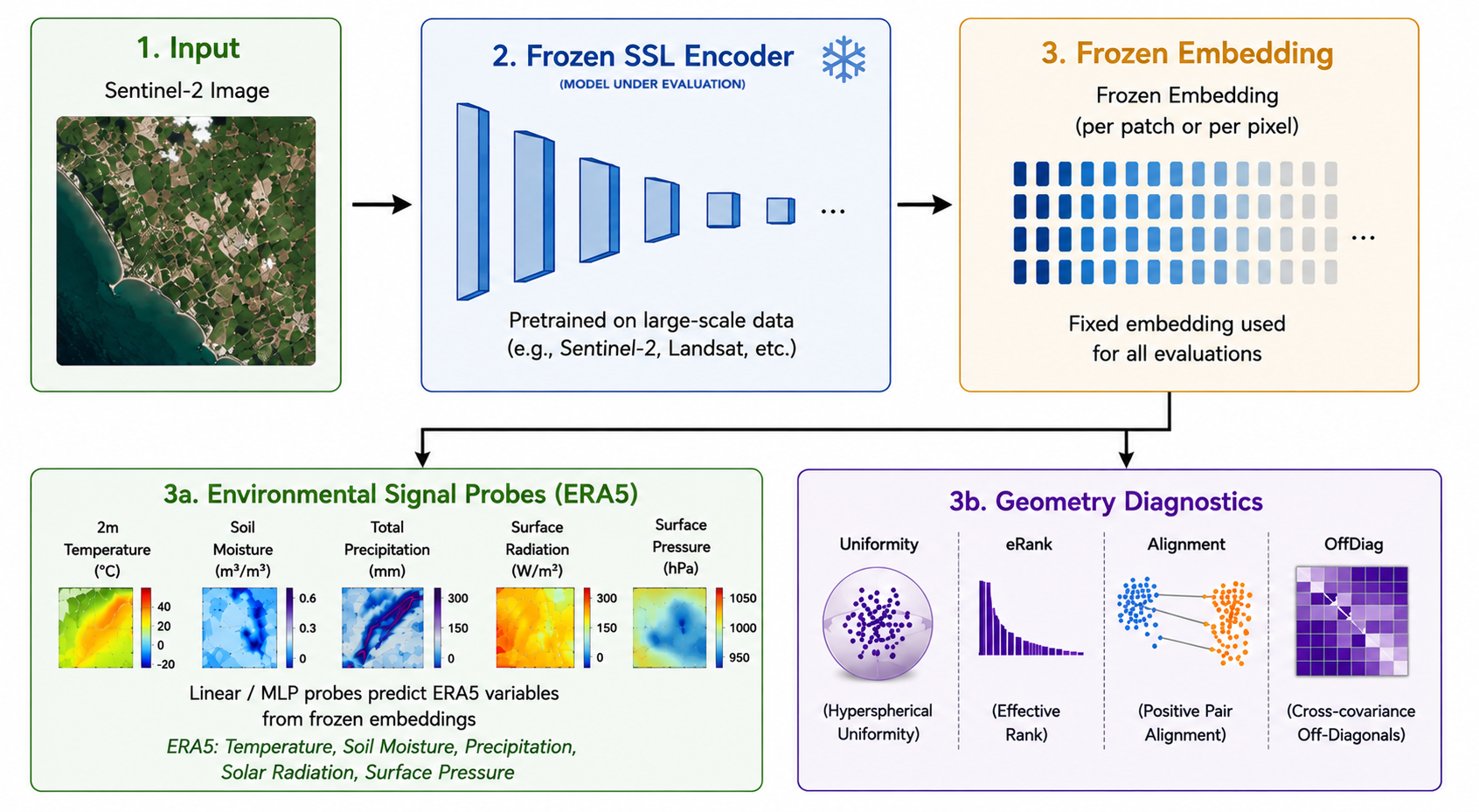}
    \caption{\textbf{Overview of the proposed evaluation framework.} Frozen encoders pretrained on large-scale Earth observation data produce fixed embeddings that are evaluated along two complementary axes: (1) ERA5~\cite{hersbach2020era5} probes measuring the linear and nonlinear accessibility of physically grounded environmental variables, and (2) intrinsic representation diagnostics (Uniformity~\cite{pmlr-v119-wang20k}, Effective Rank~\cite{erank}, Alignment~\cite{pmlr-v119-wang20k}, and Off-Diagonal Covariance~\cite{offDiag}) that characterize representation geometry. Together, these analyses quantify relationships between environmental signal accessibility, representation geometry, and downstream task performance.}
    \label{fig:flowchart}
\end{figure*}

The goal of self-supervised learning (SSL) is to learn generalizable representations that transfer across tasks rather than fitting to a specific label set. Existing geospatial benchmarks primarily evaluate learned representations through downstream task performance. While downstream evaluation measures transfer to specific tasks, it provides an incomplete view of representation quality because no benchmark can exhaustively evaluate the broad range of downstream applications SSL representations are intended to support~\cite{marks2024closerlookbenchmarkingselfsupervised}. As SSL representations are designed for unknown future tasks, representation-level diagnostics provide complementary insight beyond any fixed downstream benchmark.
 Recent work has shown that even when models achieve similar downstream performance, they can organize information differently in latent space, suggesting that the mechanisms underlying their downstream performance are functionally different~\cite{plachouras2025}. This motivates evaluation protocols that characterize representations beyond task utility.

In this work, we introduce a representation-driven evaluation protocol grounded in the physical processes underlying Earth observation imagery. Rather than measuring representations solely by task performance, we probe whether they maintain statistical associations with environmental variables that co-vary with satellite observations. 
Specifically, we use ERA5 reanalysis variables~\cite{hersbach2020era5}—a global atmospheric reanalysis dataset that combines physical weather models with observational data—to estimate environmental factors such as temperature, precipitation, surface solar radiation, surface pressure, and volumetric soil water. These physically grounded variables are used as probing targets because they directly influence the spectral reflectance and radar backscatter observed in Sentinel-2 and Sentinel-1 satellite imagery, respectively.  ERA5 probing therefore serves as a physically grounded means to evaluate what environmental signals are accessible from the learned representation.

~\cref{fig:flowchart} provides an overview of the proposed evaluation protocol. We evaluate geospatial foundation models through two complementary representation-level analyses unavailable through conventional task benchmarks. First, ERA5 probing characterizes what environmental information is encoded in the learned representation using both linear and nonlinear probes. Second, intrinsic evaluation metrics adapted from computer vision, including alignment~\cite{pmlr-v119-wang20k}, uniformity~\cite{pmlr-v119-wang20k}, effective rank (eRank)~\cite{erank}, and off-diagonal covariance (offDiag)~\cite{offDiag}, characterize the geometry and structure of the representation independently of any downstream task. Across DINO~\cite{dino}, MAE~\cite{mae}, and MoCo~\cite{moco} models trained on SSL4EO~\cite{wang2022ssl4eo}, we find that, despite similar downstream performance, the models exhibit substantially different representation geometry and encode different amounts of environmentally relevant information. Furthermore, we observe that linear accessibility of environmental signals is associated with performance on environmentally dependent downstream tasks.\newline
 Our key contributions are:
\begin{itemize}
    \item We introduce a representation-driven evaluation protocol that combines physically grounded ERA5 probing with intrinsic representation diagnostics, and demonstrate that ERA5 signal accessibility is associated with downstream performance on environmentally dependent tasks.
\item We release an extension of SSL4EO~\cite{wang2022ssl4eo} with co-located ERA5 variables for Sentinel-1 and Sentinel-2 imagery to enable future representation-level evaluation.
\end{itemize}


%% file: sec/2_related.tex
\section{Related Work}
\label{sec:related}

\subsection{Evaluating Representations}

This section expands upon representation metrics that we adapt to the geospatial setting to understand the distributional and dimensional properties of representations and the decodability of information after training. 

The distributional and dimensional structure of a representation can be characterized using metrics that require no labels. Wang and Isola~\cite{pmlr-v119-wang20k} propose alignment and uniformity as geometric characterizations of contrastive representations, showing that directly optimizing these metrics produces representations competitive with contrastive learning on downstream tasks. Because alignment and uniformity are defined as properties of the representation distribution rather than of any specific training objective, they can be applied as diagnostic metrics to any SSL method. 

Complementing these distributional properties, Roy and Vetterli~\cite{erank} introduce effective rank as a continuous approximation of matrix rank that measures how evenly information is distributed across the dimensions of a representation. Effective rank captures dimensional collapse when a model maps diverse inputs into a low-dimensional subspace of its output space. Garrido \etal~\cite{rankme} demonstrate that effective rank (RankMe) is a reliable unsupervised predictor of downstream performance across multiple datasets and domains, including speech, motivating application to the geospatial domain. Intrinsic dimension, by contrast, measures the number of degrees of freedom required to capture the local variability of the data distribution as seen through the representation. Intrinsic dimension characterizes the complexity of the learned manifold independently of the embedding dimensionality, and Rao \etal~\cite{rao2025} demonstrate its relevance for implicit neural representations in Earth observation. \newline
\indent Probing is used to characterize the decodability of information from a representation. While linear evaluation on frozen backbones is the de facto standard in computer vision, it is susceptible to memorization~\cite{alain2018}. Hewitt and Liang~\cite{hewitt2019} propose a control task framework that measures selectivity, the difference between probe performance on a real target task and on a randomly relabeled control task with the same output space, to decouple what the representation encodes from what the probe memorizes. The choice of probe complexity further characterizes how information is organized within the representation. Linear regression measures whether target variables are linearly decodable from the representation, while multi-layer perceptron (MLP) probes, as universal function approximators~\cite{hornik1989}, can in principle decode any information present in the representation regardless of how it is organized.

\subsection{Geospatial Benchmarks}

Current geospatial benchmarks evaluate representations through downstream task performance. Geo-Bench \cite{lacoste2023geobenchfoundationmodelsearth}  evaluates geospatial foundation models on diverse downstream tasks spanning multiple sensors, resolutions, and application domains. PANGAEA \cite{marsocci2025pangaeaglobalinclusivebenchmark}, builds on Geo-Bench \cite{lacoste2023geobenchfoundationmodelsearth} by broadening geographic coverage and expanding the set of downstream tasks while maintaining a focus on task utility. EarthShift\cite{doerksen2026earthshift} extends benchmark evaluation to include robustness in evaluation by curating to realistic distribution shifts across geographic regions, sensors, temporal windows, and data sources. The benchmark measures distributional robustness by comparing performance in- and out-of-distribution using paired datasets from different sources, temporal windows, geographic locations, and sensors. NeuCo-Bench\cite{neucobench} evaluates the applicability of task-agnostic Earth observation representations on a broad range of downstream tasks. ~\cref{tab:benchmark_comparison} summarizes the evaluation dimensions covered by these benchmarks. To the best of our knowledge, there has been no principled evaluation of geospatial representations beyond downstream task utility. Our work complements these benchmarks by introducing a representation-level evaluation protocol that combines intrinsic representation metrics with physically grounded environmental evaluation.

\begin{table}[t]
\centering
\setlength{\tabcolsep}{4pt}
\renewcommand{\arraystretch}{1.1}
\begin{tabular}{lcccc}
\toprule
\textbf{Benchmark} &
\rotatebox{90}{\textbf{Tasks}} &
\rotatebox{90}{\textbf{Robust}} &
\rotatebox{90}{\textbf{Intrinsic}} &
\rotatebox{90}{\textbf{Grounding}} \\
\midrule
Geo-Bench      & \checkmark &  &  &  \\
PANGAEA        & \checkmark &  &  &  \\
EarthShift     & \checkmark & \checkmark &  &  \\
NeuCo-Bench    & \checkmark &  &  &  \\
\textbf{Ours}  &            &  & \checkmark & \checkmark \\
\bottomrule
\end{tabular}
\caption{\textbf{Comparison of evaluation dimensions covered by existing geospatial benchmarks.} \textbf{Tasks}: downstream task performance; \textbf{Robust}: robustness to distribution shifts; \textbf{Intrinsic}: intrinsic representation metrics; \textbf{Grounding}: physically grounded environmental evaluation. Our framework complements existing benchmarks by introducing intrinsic representation metrics and environmental grounding.}
\label{tab:benchmark_comparison}
\end{table}

%% file: sec/3_evaluation.tex
\section{Evaluation Protocol}
\label{sec:eval}
The evaluation protocol is designed to answer the following research questions: 
\begin{enumerate}
\item  Do SSL objectives for remote sensing learn representations that encode physically meaningful environmental signals not explicitly optimized during pretraining?
\item Are intrinsic representation properties associated with the encoding and linear accessibility of physically meaningful environmental signals?
\end{enumerate}

To answer these questions, we propose two complementary components of evaluation. The first component evaluates whether physically meaningful environmental variables are encoded in the learned representations through supervised probing against co-located ERA5 reanalysis data~\cite{hersbach2020era5}. The second component evaluates intrinsic geometric and structural properties of the representation without relying on external supervision. Together, these components allow us to determine whether physically meaningful environmental signals are encoded within learned representations and whether intrinsic representation properties are associated with their linear accessibility.

\subsection{Component 1: ERA5 Physical Content Probing}

Component~1 evaluates whether environmental signals are encoded in learned representations by probing against ERA5 variables. We use both ridge regression and MLP probes to distinguish between information that is linearly accessible and information that is encoded but requires a more expressive decoder.

\paragraph{MLP Probe}
Following the information-theoretic interpretation of Pimentel \etal~\cite{pimentel}, sufficiently expressive MLP probes approximate conditional mutual information and provide evidence that information is encoded within a representation. We therefore train an MLP probe to predict co-located ERA5 variables from frozen SSL embeddings. Strong performance indicates that the representation contains information about the underlying environmental variables, even if that information is not linearly accessible.

\paragraph{Ridge Regression Probe}
Ridge regression provides a low-capacity linear probe that measures the extent to which environmental variables are directly decodable from the representation. Strong performance indicates that environmental information is readily accessible without requiring nonlinear transformations. Comparing ridge regression with the MLP probe allows us to distinguish information that is linearly accessible from information requiring more expressive decoders.

\paragraph{Probe Selectivity}

High probe performance alone does not necessarily indicate that information is encoded in the representation~\cite{alain2018,hewitt2019}. Following Hewitt and Liang~\cite{hewitt2019}, we evaluate probe selectivity to verify that probe performance reflects information present in the learned representation rather than probe memorization. The experimental protocol used to evaluate selectivity is described in Section~4.

\subsection{Component 2: Intrinsic Diagnostics}

Component~2 evaluates intrinsic representation properties without using external supervision. These metrics characterize the geometry and covariance structure of the embedding space, allowing us to investigate whether intrinsic representation properties are associated with the encoding and linear accessibility of environmental signals identified in Component~1.

\paragraph{Seasonal Alignment}

To evaluate local representation geometry, we adapt the alignment metric of Wang \etal~\cite{pmlr-v119-wang20k}. Unlike the original formulation, which measures agreement between augmented views, we define positive pairs as observations from the same geographic location acquired during different seasons. This evaluates whether representations remain consistent under seasonal variation while preserving location-specific information.

In SSL4EO \cite{wang2022ssl4eo}, each geographic location is sampled at four time points throughout the year. We define $x$ and $y$ in ~\cref{eq:alignment} and ~\cref{eq:uniformity} as representations of observations from the same location acquired during different seasons. Let $f:X\rightarrow\mathbb{R}^d$ denote the frozen encoder that maps an input satellite image to a $d$-dimensional representation. Seasonal alignment is defined by \cref{eq:alignment}.
  \begin{equation}
    \mathcal{L}_{\text{align}}(f;\,\alpha)
    = \mathbb{E}_{(x,y)\sim p_{\text{pos}}}
      \left[\|f(x)-f(y)\|_2^{\alpha}\right],\quad \alpha>0.
    \label{eq:alignment}
\end{equation}
where $p_{\text{pos}}$ denotes the distribution of positive pairs, and $\alpha$ controls the exponent of the Euclidean distance. Following Wang and Isola~\cite{pmlr-v119-wang20k}, we use $\alpha=2$. Lower values indicate stronger seasonal alignment.

\paragraph{Uniformity}
Uniformity measures how uniformly representations occupy the latent hypersphere and is defined by \cref{eq:uniformity}. 
\begin{equation}
    \mathcal{L}_{\text{unif}}(f;\,t)
    = \log\,\mathbb{E}_{(x,y)\sim p_{\text{data}}}
      \left[e^{-t\|f(x)-f(y)\|_2^{2}}\right],\quad t>0.
    \label{eq:uniformity}
\end{equation}
where $p_{\text{data}}$ denotes the distribution of image pairs sampled from the dataset, and $t$ is a temperature parameter controlling the scale of the pairwise distance penalty. Following Wang and Isola~\cite{pmlr-v119-wang20k}, we use $t=2$. Lower values indicate higher uniformity.

\paragraph{Effective Rank}

Effective rank (eRank)~\cite{erank} measures effective dimensionality of the representation by calculating singular value entropy:
\begin{equation}
    \text{eRank}(A)
    = \exp\!\left(-\sum_i p_i \log p_i\right),\quad
    p_i = \frac{\sigma_i}{\sum_j \sigma_j},
    \label{eq:erank}
\end{equation}
where $A \in \mathbb{R}^{N \times d}$ is a matrix of $N$ representations of dimension $d$. $\{\sigma_i\}$ are the singular values of the matrix $A$ and $p_i$ is the normalized singular value distribution. Higher eRank indicates information spread across many embedding dimensions.

\paragraph{Off Diagonal Covariance}

Off-diagonal covariance~\cite{offDiag} measures representational redundancy by quantifying the correlation between embedding dimensions using the energy of the off-diagonal entries of the representation covariance matrix, as defined in Equation~\ref{eq:offdiag}.
\begin{equation}
    \text{offDiag}(Z)
    = \sum_{i \neq j} \bigl[C(Z)\bigr]_{ij}^{2},
    \label{eq:offdiag}
\end{equation}
where $C(Z)$ is the empirical covariance matrix computed from a batch of representations $Z$. The summation is taken over the off diagonal entries $(i \neq j)$ Higher off-diagonal covariance indicates greater representational redundancy.

%% file: sec/4_experiment.tex
\section{Experimental Setup}

We evaluate the proposed protocol in two settings. First, we perform a controlled study using DINO~\cite{dino}, MAE~\cite{mae}, and MoCo~\cite{moco} models trained under identical conditions on SSL4EO~\cite{wang2022ssl4eo}. Second, we evaluate publicly available geospatial foundation models to investigate whether the observed relationships generalize across architectures, pretraining datasets, and model scales. Throughout all experiments, ERA5 variables serve as the probing targets because they are associated with environmental processes reflected in satellite imagery while remaining absent from the SSL pretraining objectives.

\subsection{ERA5 Extension of SSL4EO}

We extend the SSL4EO~\cite{wang2022ssl4eo} dataset by associating each Sentinel-1 and Sentinel-2 observation with co-located ERA5 reanalysis variables~\cite{hersbach2020era5} corresponding to the image acquisition date. For every image, we extract daily aggregates of surface temperature, total precipitation, surface solar radiation, surface pressure, and volumetric soil water. Summary statistics for both sensing modalities, quality-control procedures, and spatial and temporal coverage analyses are provided in ~\cref{app:curation}.

To reduce probe training cost while preserving the geographic distribution of the original dataset, we construct a spatially balanced training subset using kernel density estimation (KDE). Probe training is performed on the downsampled training split, while all evaluation is conducted on the held-out SSL4EO \cite{wang2022ssl4eo} validation split. The subset size was selected through saturation experiments that measured probe performance as a function of training set size, as shown in \cref{app:hyperparameter}.

\subsection{Link between ERA5 and Sentinel Imaging}

Sentinel-1 and Sentinel-2 record radar backscatter and spectral reflectance, respectively, both of which are influenced by surface and atmospheric conditions. These physical relationships motivate the design of Earth observation sensors and the selection of spectral wavelengths, resulting in statistical associations between environmental variables and the recorded satellite observations. We focus on temperature, precipitation, surface solar radiation, surface pressure, and volumetric soil water because they characterize complementary components of the land-atmosphere system that influence vegetation dynamics, surface moisture, and atmospheric conditions reflected in satellite imagery. ERA5 probing is therefore used as a physically grounded diagnostic of environmental information encoded within the learned representation rather than as a measure of downstream task performance or a causal model of sensor physics.

\subsection{SSL4EO Control}

We evaluate ViT-S/16 models trained under the SSL4EO protocol using DINO~\cite{dino}, MAE~\cite{mae}, and MoCo~\cite{moco} in a controlled experimental setting. All models are trained for 100 epochs using the 13-band Sentinel-2 L1C imagery provided by SSL4EO. Representations are extracted from the final layer of each frozen encoder and evaluated using the protocol described in Section \ref{sec:eval}.

Linear ridge regression and MLP probes are trained on the downsampled 50K SSL4EO~\cite{wang2022ssl4eo} training split and evaluated on the held-out validation split. The MLP probe consists of a two-layer multilayer perceptron with a hidden dimension of 256, ReLU activations, and a dropout rate of 0.1. For precipitation, we predict the logarithm of accumulated precipitation to account for the highly skewed target distribution. Probe performance is reported as the average coefficient of determination ($R^2$) across the five ERA5 variables.

For comparison between the SSL4EO~\cite{wang2022ssl4eo} control models and geospatial foundation models, ridge regression uses a fixed regularization parameter of $\alpha=10{,}000$ for all models to ensure a consistent evaluation protocol. For the selectivity analysis, the ridge regression regularization parameter is optimized independently for each SSL objective using a validation sweep to ensure that each method is evaluated under its best-performing setting. 

Following Hewitt and Liang~\cite{hewitt2019}, we evaluate probe selectivity by repeating probe training using randomly permuted ERA5 targets while keeping image embeddings fixed. The difference between performance on the true and randomized targets indicates whether the probe is recovering information encoded in the representation rather than memorizing the regression task. Each probe is trained three times with different random seeds, and the mean performance is reported.

\begin{figure}[t]
\centering
\includegraphics[width=0.45\textwidth]{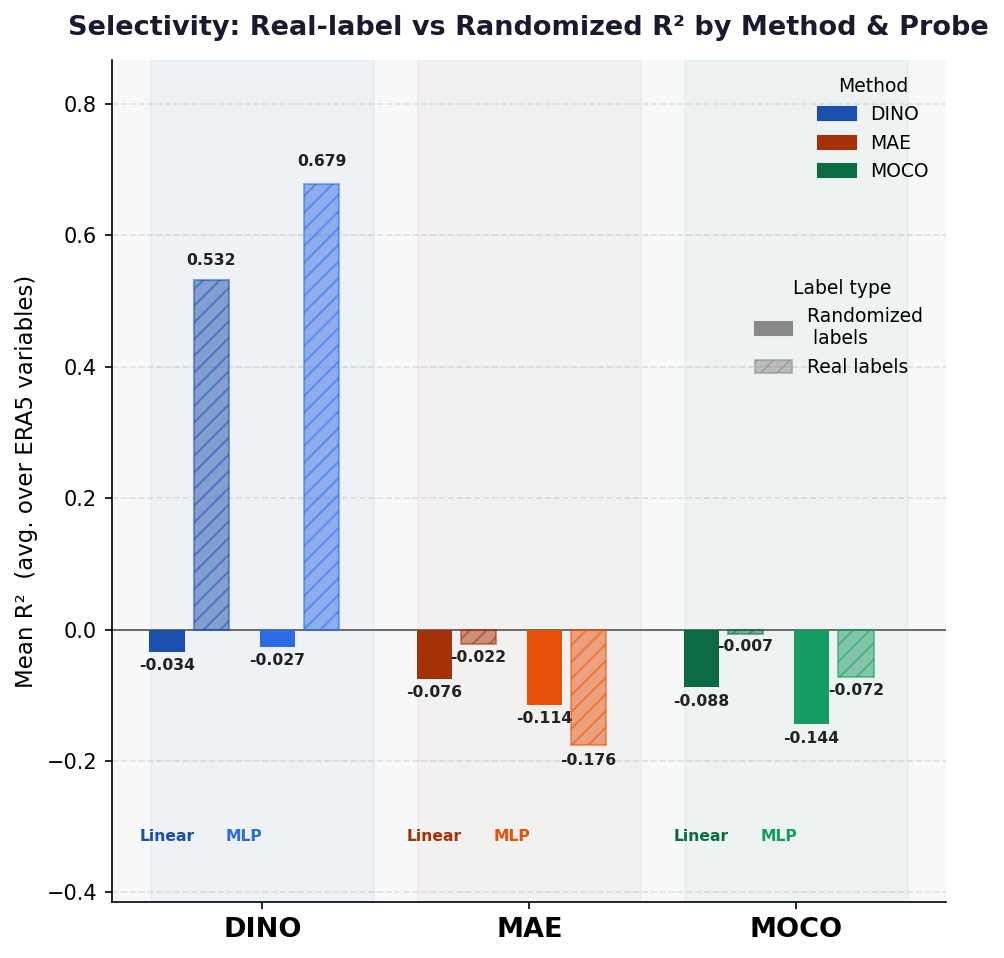}
\caption{\textbf{Probe selectivity analysis for the control models.} Both the linear and MLP probes exhibit high selectivity for DINO, indicating that ERA5 environmental information is encoded in the representation. MAE~\cite{mae} and MoCo~\cite{moco} show slight improvement on ground-truth targets compared to randomized targets.}
\label{fig:selectivity}
\end{figure}

\input{sec/diagnostic_tables}

\begin{figure*}[t]
\centering
\includegraphics[width=0.85\textwidth]{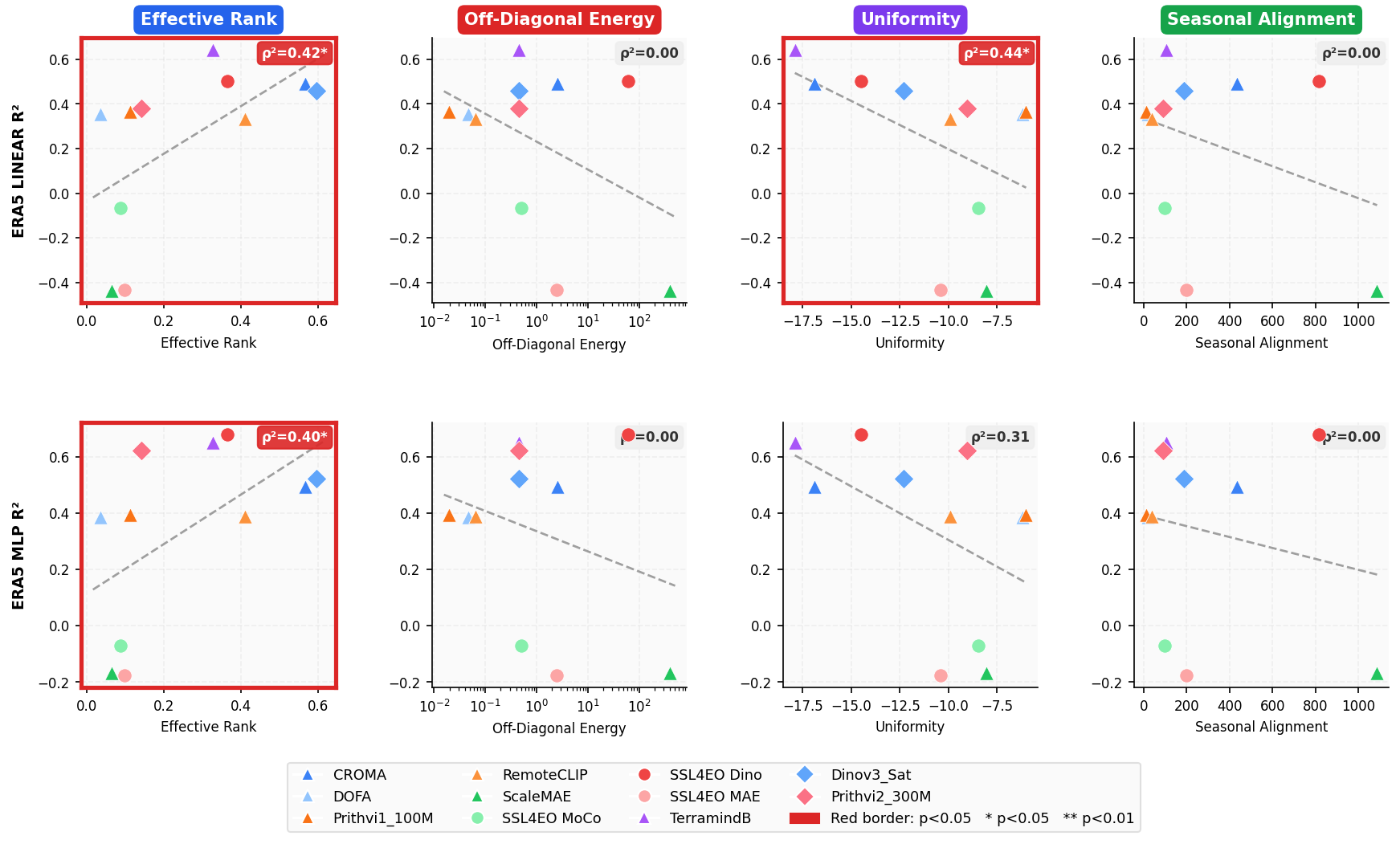}
\caption{\textbf{Relationship between diagnostic metrics and probe performance.} To identify the relationship between diagnostic metrics and probe performance we calculate the Spearman coefficient for the evaluated models. Metrics that show a relationship to ERA5 probe performance are outlined in red. We observe an association between eRank and the linear ($\rho_{linear}=0.42$) and MLP probe ($\rho_{MLP}=0.40$). We observe a moderate association ($\rho_{linear}=0.44$) between uniformity and linear probe performance as both measure linear separability in different settings.}
\label{fig:diagnostics}
\end{figure*}

\begin{figure*}[t]
\centering
\includegraphics[width=0.85\textwidth]{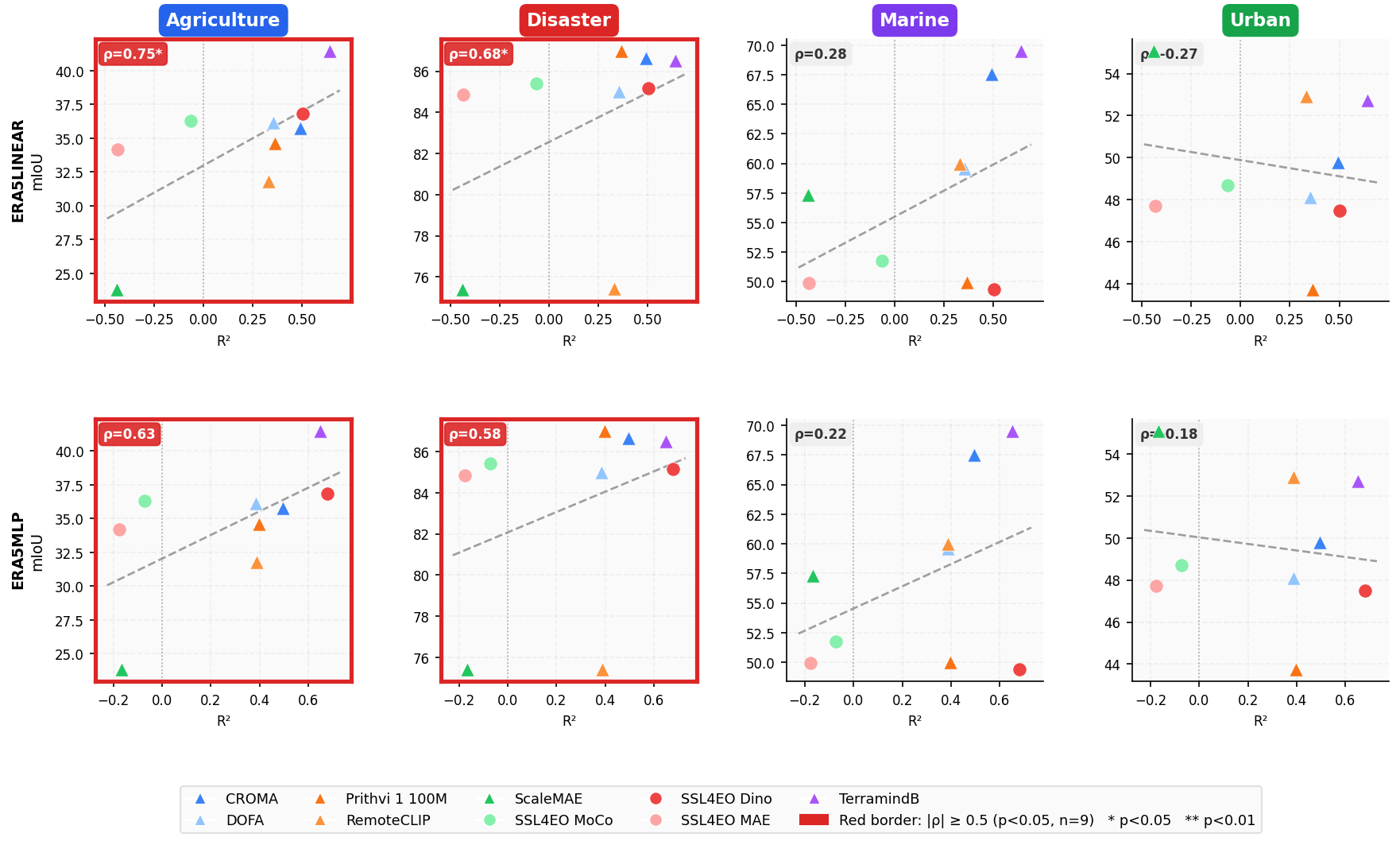}
\caption{
\textbf{Relationship between ERA5 probe performance and downstream task performance across PANGAEA~\cite{marsocci2025pangaeaglobalinclusivebenchmark} application domains.} Domain performance is measured as the mean mIoU across datasets within each domain: agriculture (PATIS~\cite{garnot2021multimodaltemporalattentionmodels,garnot2022panopticsegmentationsatelliteimage}, Crop Type Mapping~\cite{Rustowicz2019SemanticSO,yeh2021sustainbenchbenchmarksmonitoringsustainable}, AI4Farms~\cite{persello2023ai4smallfarms}), disaster (Sen1Fl11~\cite{rambour2020flood}, HLS BurnScar~\cite{prithviv1}), marine (MADOS~\cite{kikaki2024detecting}), and urban (Five Billion Pixels~\cite{tong2023enabling}, DynamicEarthNet~\cite{toker2022dynamicearthnet}, SpaceNet~7~\cite{vanetten2019spacenet}). The Spearman coefficient quantifies the association between ERA5 probe and downstream performance. Associations are strongest for agriculture and disaster tasks, with linear probes outperforming MLP probes (agriculture: $\rho_{\text{linear}}=0.75$ vs. $\rho_{\text{MLP}}=0.63$; disaster: $\rho_{\text{linear}}=0.68$ vs. $\rho_{\text{MLP}}=0.58$).}
  \label{fig:eravtask}
\end{figure*}
\begin{figure}[t]
\centering
\includegraphics[width=0.45\textwidth]{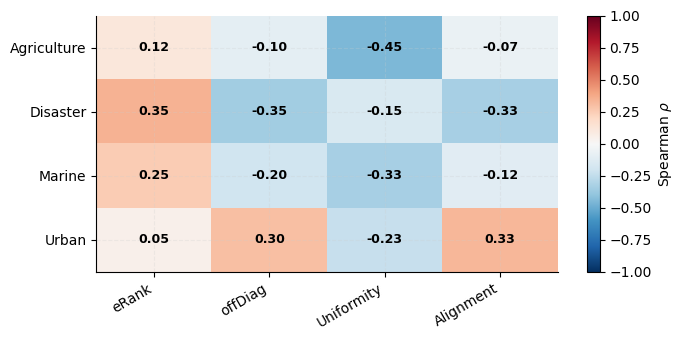}
\caption{\textbf{Relationship between task performance and diagnostic metrics.} We use the same task breakdown as Figure \ref{fig:eravtask} and find the Spearman correlation to diagnostic metrics. Uniformity shows strongest correlation to agriculture downstream task performance ($\rho_{unif}=-0.45$). Other metrics have moderate to weak association to downstream tasks}
\label{fig:geovtask}
\end{figure}

\subsection{Geospatial Foundation Models}

We evaluate publicly available geospatial foundation models using the same protocol to determine whether relationships observed in the controlled setting generalize beyond SSL4EO \cite{wang2022ssl4eo}. Inputs are preprocessed using the normalization statistics and spectral bands used during each model's pretraining. For consistency, all models are evaluated using Sentinel-2 imagery only, with additional modalities and metadata available during pretraining omitted at inference. When models produce token-level outputs, we use the CLS token when available and otherwise obtain a single representation through mean pooling of patch embeddings. All extracted representations are evaluated using the probing and intrinsic diagnostic protocol described in Section \ref{sec:eval}.

%% file: sec/diagnostic_tables.tex
%

\begin{table*}[t]
  \centering
  \label{tab:diagnostics}
  \begin{tabular}{lcccccc}
    \toprule
    & \multicolumn{2}{c}{\textbf{Environmental Grounding}} & \multicolumn{4}{c}{\textbf{Intrinsic Representation Metrics}} \\
    \cmidrule(r){2-3} \cmidrule(l){4-7}
    \textbf{Model} &
    \textbf{Linear $\mathbf{R^2}$} $\uparrow$ &
    \textbf{MLP $\mathbf{R^2}$} $\uparrow$ &
    \textbf{eRank} $\uparrow$ &
    \textbf{offDiag.} $\downarrow$ &
    \textbf{Unif.} $\downarrow$ &
    \textbf{Align.} $\downarrow$ \\
    \midrule
    CROMA~\cite{croma} & 0.49 & 0.49 & 0.57 & 2.60 & -16.90 & 433.48 \\
    DOFA~\cite{dofa} & 0.35 & 0.39 & 0.04 & 0.05 & -6.19 & 20.78 \\
    PrithviV1 100M~\cite{prithviv1} & 0.36 & 0.39 & 0.11 & \textbf{0.02} & -6.03 & \textbf{10.48} \\
    RemoteCLIP~\cite{remoteclip} & 0.33 & 0.39 & 0.41 & 0.07 & -9.92 & 36.81 \\
    ScaleMAE~\cite{scalemae} & -0.44 & -0.17 & 0.06 & 401.76 & -8.05 & 1086.16 \\
    SSL4EO MoCo~\cite{moco,wang2022ssl4eo} & -0.07 & -0.07 & 0.09 & 0.50 & -8.45 & 97.69 \\
    SSL4EO Dino~\cite{dino,wang2022ssl4eo} & 0.50 & \textbf{0.68} & 0.36 & 61.42 & -14.52 & 817.28 \\
    SSL4EO MAE~\cite{mae,wang2022ssl4eo} & -0.43 & -0.18 & 0.10 & 2.51 & -10.41 & 197.12 \\
    Terramind B~\cite{terramind} & \textbf{0.64} & 0.65 & 0.33 & 0.45 & \textbf{-17.88} & 105.15 \\
    DINOv3 Sat~\cite{dinov3} & 0.46 & 0.52 & \textbf{0.60} & 0.45 & -12.33 & 186.16 \\
    PrithviV2 300M~\cite{prithvieo2} & 0.38 & 0.62 & 0.14 & 0.45 & -9.06 & 89.45 \\
    \bottomrule
  \end{tabular}
  \caption{\textbf{Control and geospatial foundation model performance.} We evaluate the performance of SSL4EO control models and standard geospatial foundation models against the ERA5 probing and diagnostic metric evaluation protocol. The intrinsic diagnostics include effective rank (eRank), off-diagonal covariance (offDiag.), uniformity (Unif.), and seasonal alignment (Align.). Physical content is evaluated using the average coefficient of determination ($R^2$) across ERA5 variables for both a linear probe (Linear $R^2$) and a MLP probe (MLP $R^2$). Effective rank and off-diagonal covariance are normalized by embedding dimensionality to enable comparison across models with different embedding sizes. \textbf{Bold} refers to best performing model for that metric.}
  \label{tab:results}
\end{table*}

%% file: sec/5_results.tex
\section{Results and Discussion}
\label{sec:results}

We evaluate the SSL4EO \cite{wang2022ssl4eo} control models and a collection of geospatial foundation models using the proposed evaluation protocol. In the controlled setting, DINO shows the strongest decodability of ERA5 variables, while MAE and MoCo remain closer to the randomized baseline. Across the broader model set, representation geometry varies substantially, and the relationship between representation geometry, ERA5 decodability, and downstream performance depends on the task domain. Table \ref{tab:results} shows the complete results. 

\paragraph{SSL representations encode physically meaningful environmental signals.}
 In the control setting, SSL4EO  \cite{wang2022ssl4eo} checkpoints trained on the identical data encode different amounts of environmental signals. The selectivity experiment in Figure \ref{fig:selectivity} shows that probe performance on true ERA5 labels consistently exceeds the randomized baseline, with DINO exhibiting the largest gap, providing the strongest evidence of a decodable environmental signal. Although the three SSL4EO models  \cite{wang2022ssl4eo} achieve similar downstream mIoU on PANGEA  \cite{marsocci2025pangaeaglobalinclusivebenchmark}, they differ substantially in how much ERA5 signal that can be decoded from their representations. A similar trend is observed in geospatial foundation models, notably, that DINOv3 Sat~\cite{dinov3} achieves high MLP probe results for environmental variables despite RGB-only input, outperforming DOFA~\cite{dofa} and CROMA~\cite{croma}, both of which were pretrained with multispectral images with more direct access to ERA5-relevant information.
 This suggests that physically meaningful signals can be recovered from geospatial foundation models even when the input modality does not directly observe all of the environmental variables considered in this work. Because the foundation models differ in architecture, parameter count, and pretraining data, we cannot attribute these differences to any individual design choice.

\paragraph{Some representation properties are associated with environmental signal encoding.}

Normalized effective rank shows a positive association with both linear ($\rho=0.42$) and MLP ($\rho=0.40$) probe performance (\cref{fig:diagnostics}). Uniformity is associated with linear probe performance ($\rho=-0.45$), which is consistent with both metrics capturing aspects of linear accessibility. In contrast, offDiag~\cite{offDiag} and seasonal alignment~\cite{pmlr-v119-wang20k} show little association with the ability to decode environmental signals (\cref{fig:diagnostics}). These results suggest that not all intrinsic representation diagnostics are equally informative for characterizing environmental signal encoding.

\paragraph{Environmental signal encoding is most strongly associated with agriculture and disaster tasks.}
While PANGAEA reports overall benchmark performance averaged across all tasks, the benchmark also categorizes datasets into agriculture, disaster, marine, and urban application domains. We leverage this taxonomy to investigate whether the relationship between ERA5 probe performance and downstream performance differs across domains. We calculate the Spearman correlation between diagnostic metrics, probe regression, and PANGAEA benchmark \cite{marsocci2025pangaeaglobalinclusivebenchmark} broken down into domain.

ERA5 probes show the strongest relationship with agriculture and disaster task performance (\cref{fig:eravtask}). Notably, the linear probe shows a stronger relationship than the MLP probe, suggesting that the ease with which ERA5 information can be decoded is more indicative of downstream task utility than the mere presence of that information in the representation. Among the intrinsic metrics, uniformity shows the strongest association with agriculture task performance (Figure~\ref{fig:geovtask}). Lower uniformity is associated with higher linear separability, and lower uniformity scores are associated with high agriculture task performance. This suggests that, for frozen-backbone models on agriculture tasks, linear accessibility of environmental variables may be more important than only their presence in the representation. Although both uniformity and linear probing capture aspects of linear accessibility, Table~\ref{tab:results} suggests that geometry alone cannot explain ERA5 probe performance. Models like SSL4EO MAE~\cite{mae,wang2022ssl4eo}, RemoteCLIP~\cite{remoteclip}, and PrithviV2 300M~\cite{prithvieo2} show similar uniformity scores, but substantially different ERA5 linear $R^2$ values. This suggests that uniformity may be a geometric property associated with strong performance on agriculture tasks, but it is not sufficient on its own.

\paragraph{Representation metrics show limited association with out-of-distribution robustness.}
Finally, we investigate the relationship between the representation properties for models and performance on the EarthShift \cite{doerksen2026earthshift} robustness benchmark. We observe weak associations between alignment and uniformity and robustness under geographic, sensor, and temporal distribution shifts. In contrast, the ERA5 probe metrics, off-diagonal energy~\cite{offDiag}, and effective rank~\cite{erank} show little to no association with out-of-distribution performance. Given the relatively small number of models evaluated, these results should be viewed as an initial exploratory analysis rather than evidence of a general relationship. Complete Spearman correlation results between the representation metrics and EarthShift \cite{doerksen2026earthshift} distribution shifts are provided in ~\cref{app:ood_analysis}.

%% file: sec/6_conclusion.tex
\section{Conclusion and Limitations}
\label{sec:conclusion}

Our results demonstrate that ERA5 decodability provides a meaningful axis for evaluating self-supervised representations in remote sensing, complementing purely task-based benchmarks by grounding model evaluation in the physical world. Across both controlled SSL4EO models and geospatial foundation models, we find that models with similar downstream performance can encode substantially different amounts of environmentally relevant information. Furthermore, we find that the linear decodability of environmental variables, rather than their mere presence, is associated with downstream task utility, and intrinsic representation properties, specifically uniformity, are associated with structure relevant to both environmental encoding and performance on environmentally dependent tasks. Geometry alone is insufficient to explain environmental signal encoding. 

While our analysis is associative rather than causal, and ERA5 variables represent one of many possible grounding signals, we believe that linking representations to physically meaningful quantities offers a promising direction for understanding what general-purpose remote sensing models learn. Future work should expand this evaluation to a larger set of models and diagnostic metrics, and explore whether richer environmental grounding signals further explain downstream performance and out-of-distribution robustness. More broadly, our results suggest that representation-level analyses provide complementary information to downstream benchmarks by characterizing properties of learned representations that are not directly observable from task performance alone. 

\paragraph{LLM Usage}
ChatGPT was used for editorial assistance and draft conceptual illustrations. Claude Code was used for code generation to produce tables and figures. All scientific content, experiments, analyses, and conclusions are the authors' own.

\FloatBarrier

%% file: sec/appendix.tex
\section{Appendix}
\subsection{Dataset Curation  and Statistics}
\label{app:curation}
\begin{figure}[ht]
  \centering
  \includegraphics[width=.8\columnwidth]{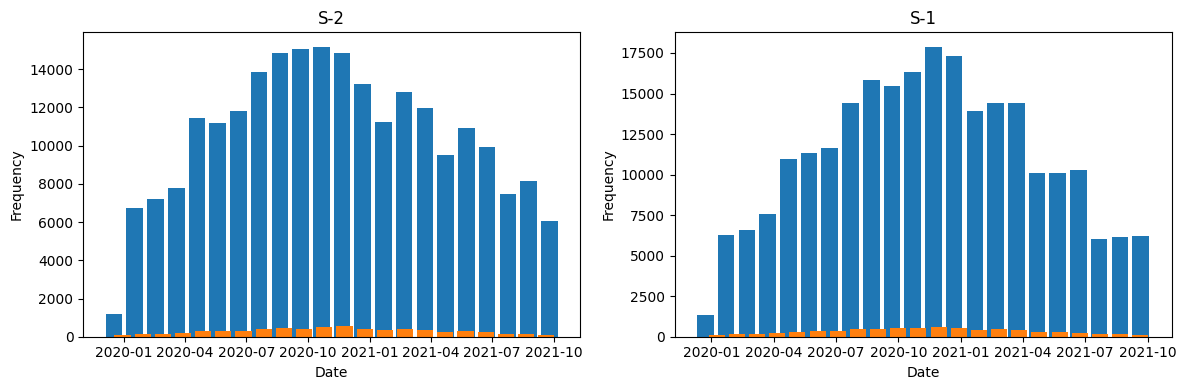}
  \caption{\textbf{Temporal distribution of ERA5 variables for S-1 and S-2.} The two figures show the temporal distribution for train (blue) and test (orange) subsets.}
  \label{fig:tempstatss2}
\end{figure}

\begin{figure}[t]
  \centering
  \includegraphics[width=.8\columnwidth]{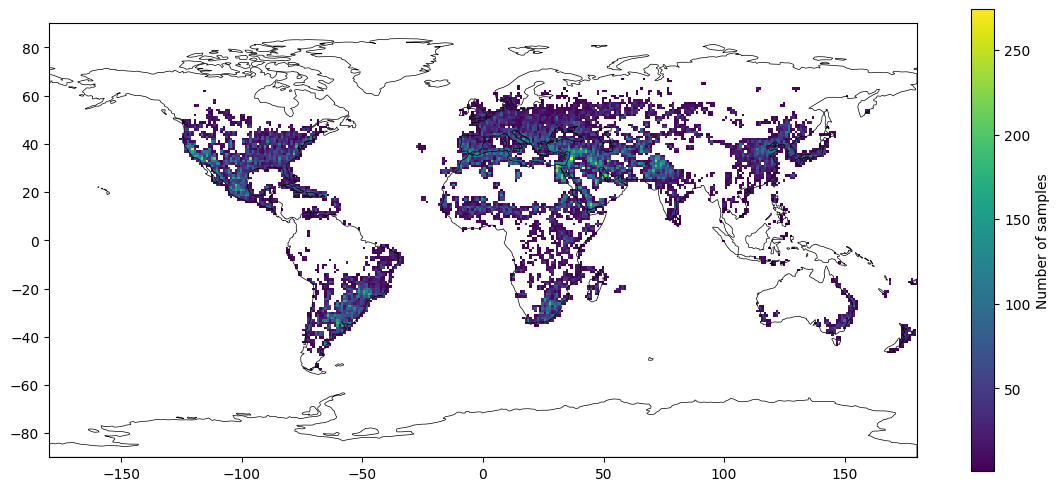}
  \caption{\textbf{Spatial distribution of ERA5 variables for S-2 and S-1.} The spatial graph shows both the training and test set combined, showcasing the global coverage of the dataset}
  \label{fig:spatial_viz}
\end{figure}

\begin{figure}[t]
  \centering
  \includegraphics[width=.9\columnwidth]{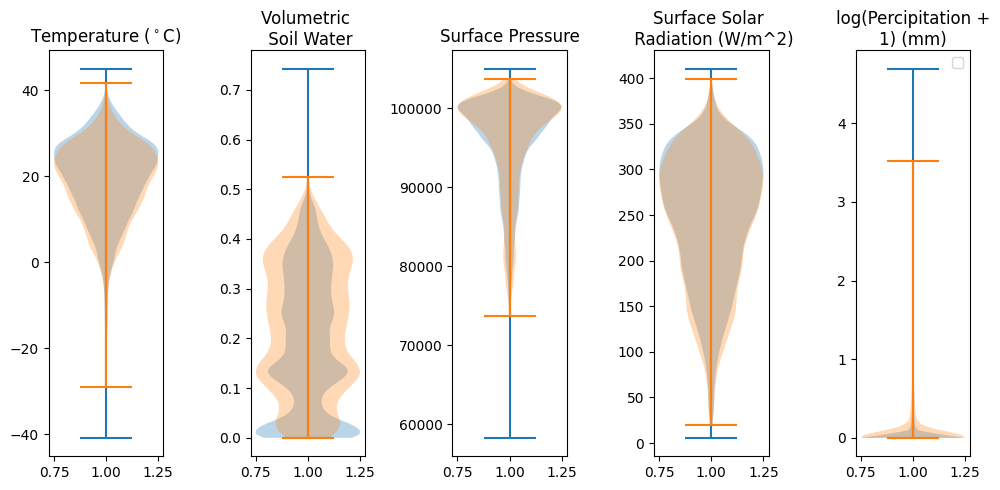}
  \includegraphics[width=.9\columnwidth]{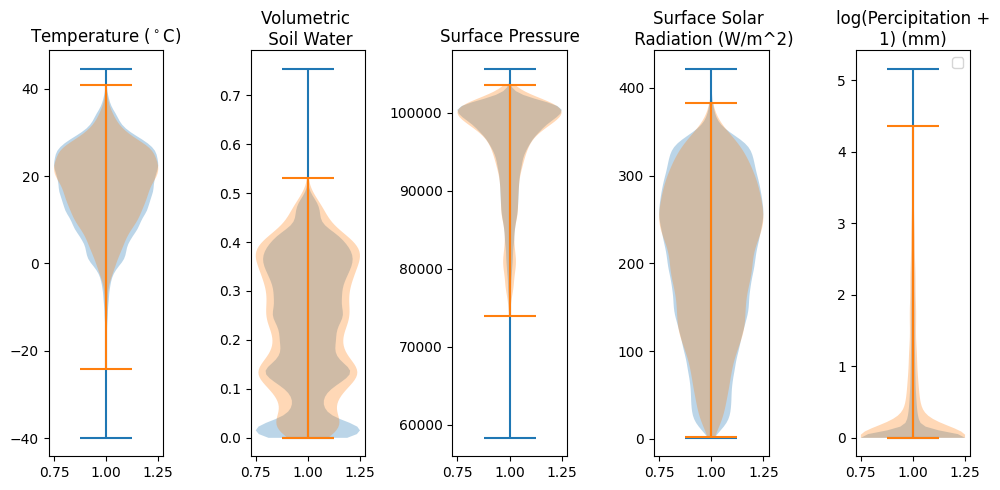}
  \caption{\textbf{Distribution of ERA5 variables for split training and test sets for S2 and S1}. The two figures show the distribution of each of the ERA5 variables for train (blue) and test (orange) subsets. The figure on the top is the S-2 co-locate ERA5 distribution and the figure on the bottom is the S-1 co-locate ERA5 variables.}
  \label{fig:statss2}
\end{figure}

\begin{figure*}[t]
\centering
\includegraphics[width=0.7\textwidth]{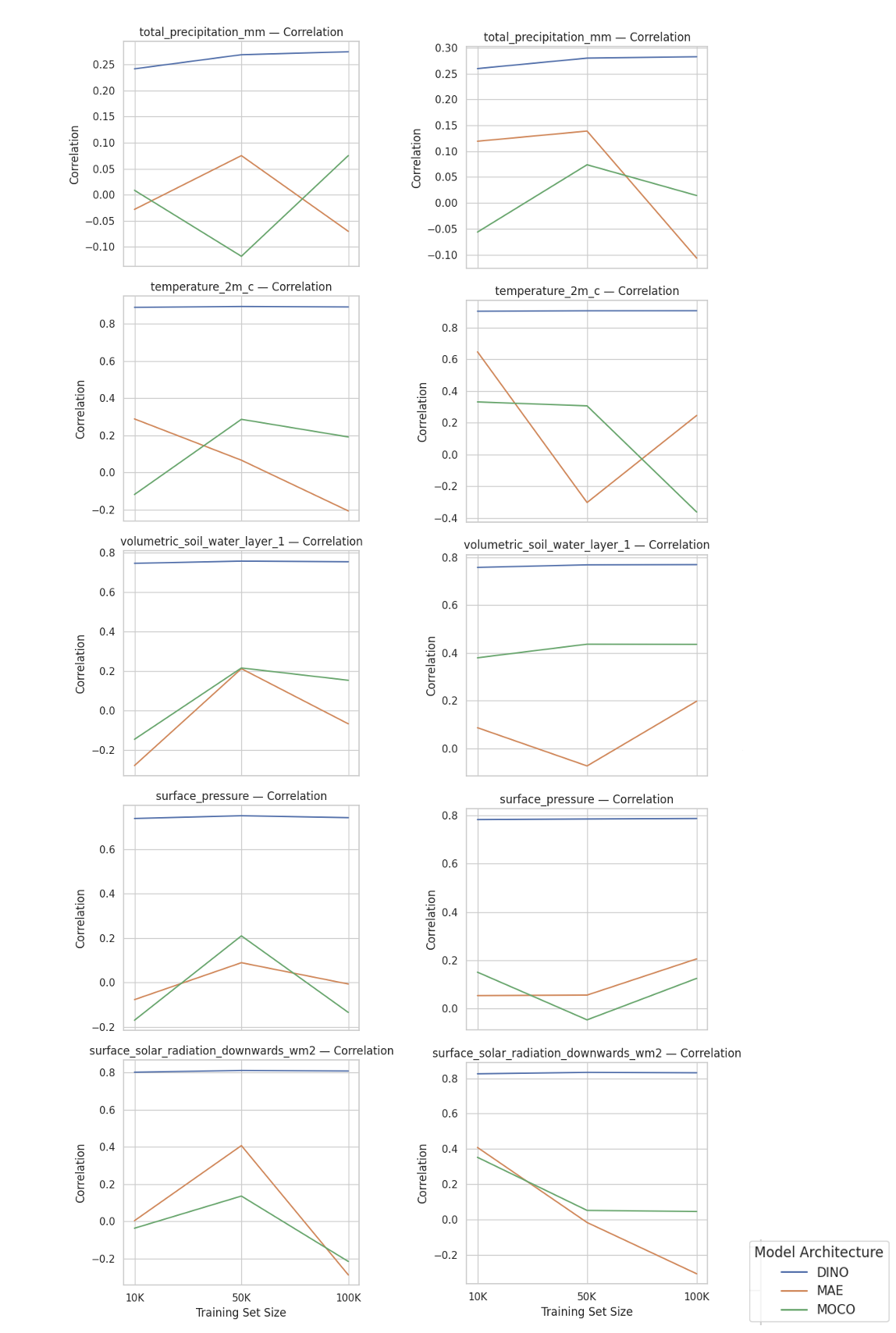}
\caption{\textbf{Probe performance as a function of training set size.} Saturation experiment used to determine the probe training subset size. Ridge regression (left) exhibits little improvement beyond approximately 50K training samples. The MLP probe (right) is trained using a fixed optimization budget of 8K gradient steps; consequently, larger training sets receive fewer effective training epochs. Based on these results, we use a 50K training subset for all subsequent probing experiments as a computationally efficient operating point.}
\label{fig:sample_size}
\end{figure*}

This section summarizes the statistics of the co-located ERA5 variables used throughout the study. The dataset includes surface solar radiation, 2 m temperature, surface pressure, volumetric soil water (layer 1), and total precipitation. \Cref{tab:era5stats} reports summary statistics for the ERA5 variables associated with both Sentinel-1 (S-1) and Sentinel-2 (S-2) observations. Because the two sensors are acquired in close temporal proximity, the corresponding ERA5 distributions are nearly identical across modalities. \Cref{tab:era5qc} lists the quality-control thresholds used to remove physically implausible or extreme values. \Cref{fig:tempstatss2,fig:spatial_viz} illustrate the temporal and spatial distributions of the curated dataset, while \Cref{fig:statss2} shows the distributions of each ERA5 variable for the training and validation splits of both sensing modalities. Finally, \Cref{fig:sample_size} presents the saturation experiment used to determine the probe training subset size.

\begin{table}[ht]
\centering
\small
\begin{tabular}{llrr}
\toprule
Variable & Stat & S-2 & S-1 \\
\midrule
Surface Solar Radiation & Mean & 235.1 & 235.1 \\
(W/m\textsuperscript{2}) & Std & 77.8 & 77.8 \\
& Cardinality & 867164    & 867216  \\
Temperature & Mean & 17.7 & 17.87 \\
(\textdegree C) & Std & 9.9 & 9.8 \\
& Cardinality & 867164 & 867216 \\
Surface Pressure & Mean & 95998 & 95997 \\
(Pa) & Std & 6058 & 6058 \\
& Cardinality & 867164 & 867216 \\
Volumetric Water & Mean & 0.23 & 0.22 \\
(m\textsuperscript{3}/m\textsuperscript{-3}) & Std & 0.13 & 0.13 \\
 & Cardinality & 867164 & 867216 \\
Precipitation & Mean & 0.38 & 0.38 \\
(mm) & Std & 1.79 & 1.79 \\
& Cardinality & 867164 & 867216 \\
\bottomrule
\end{tabular}
\caption{\textbf{Summary statistics of SSL4EO co-located ERA5 variables.} Mean and standard deviation are computed over all valid image–ERA5 pairs after quality filtering. Cardinality denotes the number of co-located observations for each variable.}
\label{tab:era5stats}
\end{table}

\begin{table}[t]
\centering
\small
\begin{tabular}{lrr}
\toprule
Variable & Min & Max \\
\midrule
Surface Solar Radiation & 0 & 463.0 \\
(W/m\textsuperscript{2}) &  &  \\
Temperature & -93.15 & 66.85  \\
(\textdegree C) &  &   \\
Surface Pressure & $5e4$ &  $1.2e5$ \\
(Pa) &  &  \\
Volumetric Water & 0 & 1.0  \\
(m\textsuperscript{3}/m\textsuperscript{-3}) &  &  \\
Precipitation & 0 & 1,000 \\
(mm) &  &  \\
\bottomrule
\end{tabular}
\caption{\textbf{ERA5 Quality control thresholds.} To control for extreme values or physically implausible values each ERA5 variable was contained within the range outlined}
\label{tab:era5qc}
\end{table}

\clearpage

\FloatBarrier

\subsection{Probe Hyperparameter}
\label{app:hyperparameter}
\begin{figure*}[t]
  \centering
  \includegraphics[width=0.75\textwidth]{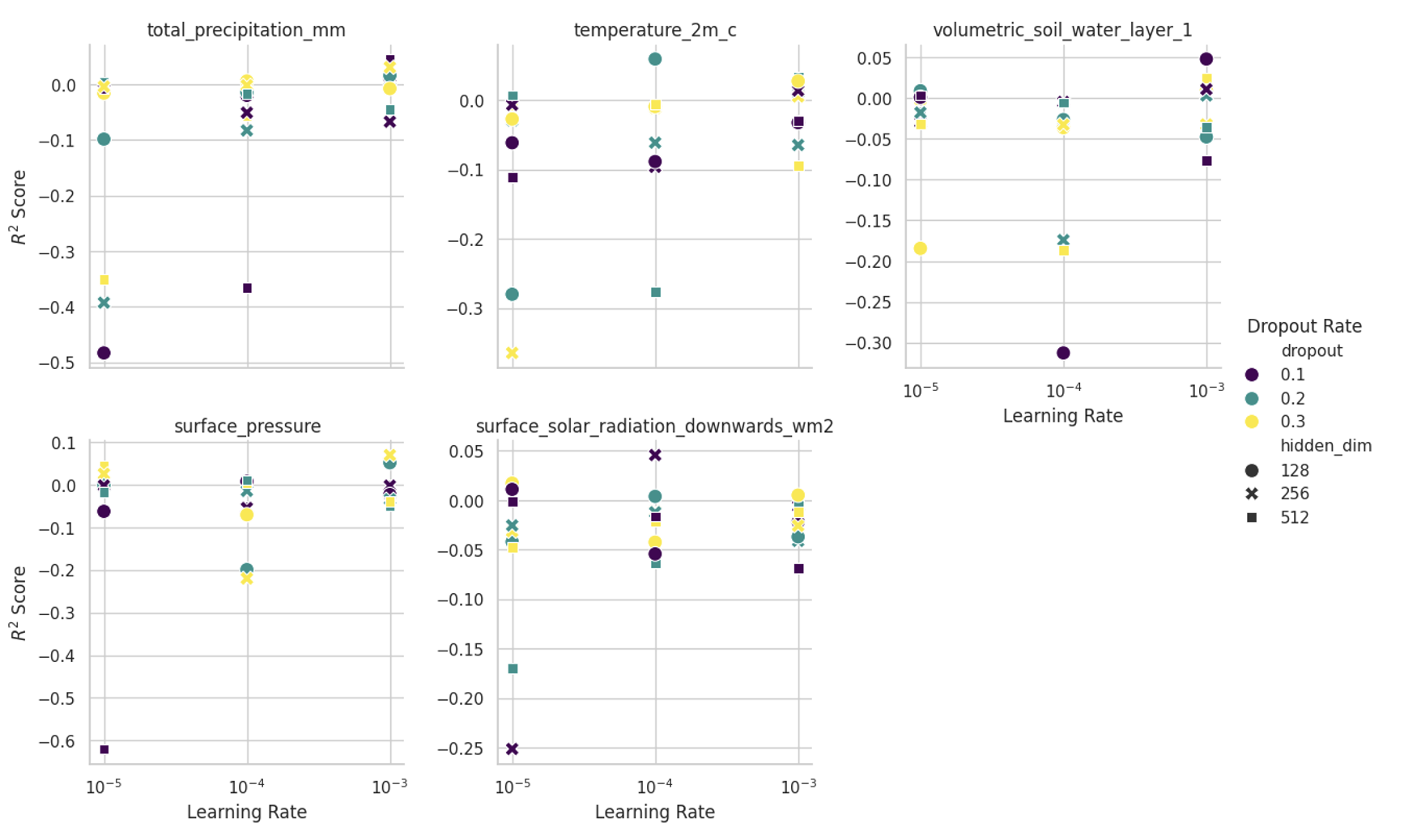}
  \includegraphics[width=0.75\textwidth]{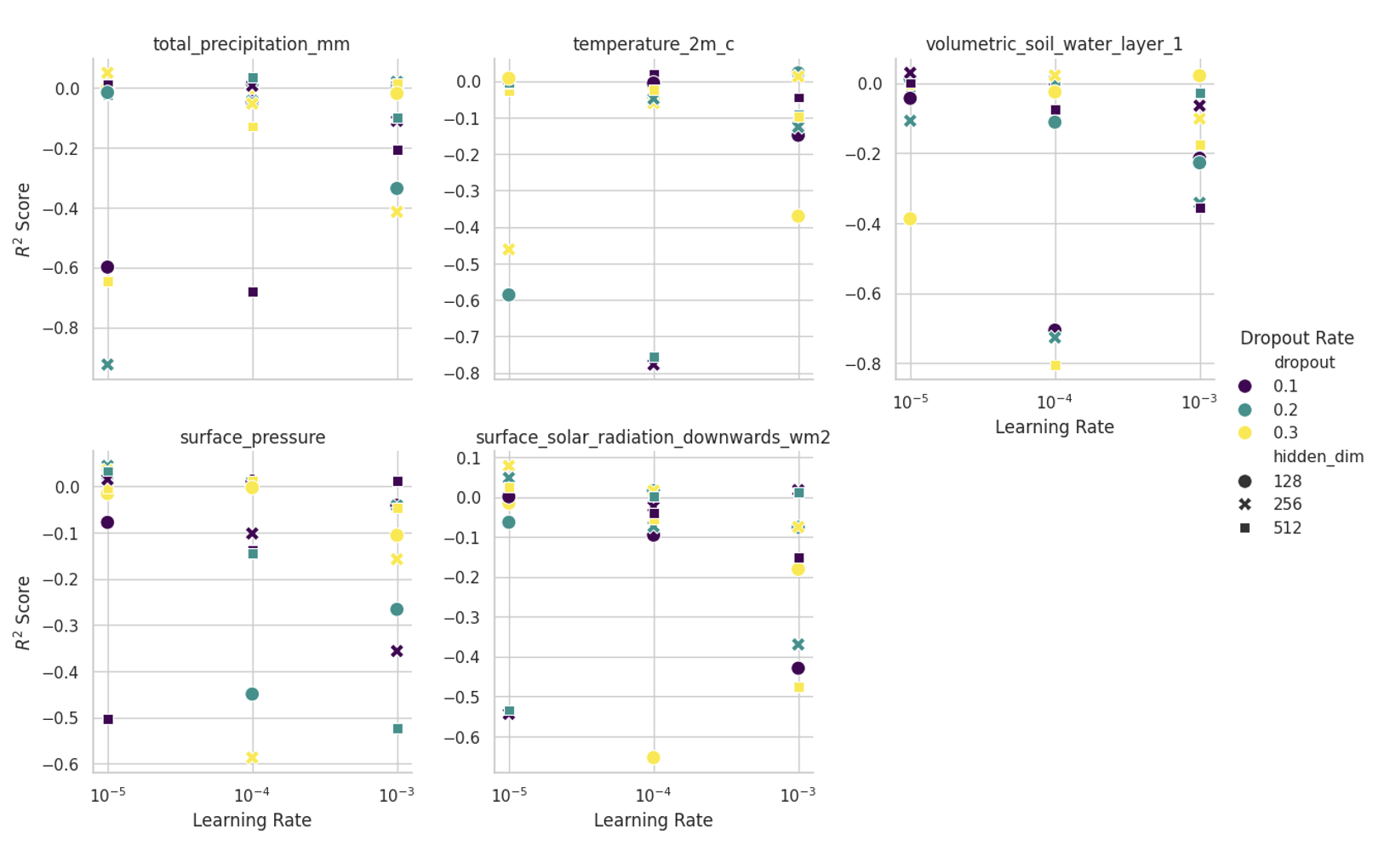}

  \caption{\textbf{MLP hyperparameter search.} We perform a grid search over learning rate, hidden dimension, and dropout rate. The selected configuration corresponds to the highest mean validation $R^2$ across the five ERA5 prediction tasks. The top graph shows the results for the MoCo objective, while the bottom graph shows the results for the MAE objective. The DINO objective is omitted because it showed minimal variation across the evaluated hyperparameter configurations.}
    \label{fig:mlp_probe}
\end{figure*}

\begin{figure*}[t] 
  \centering
  \includegraphics[width=0.75\textwidth]{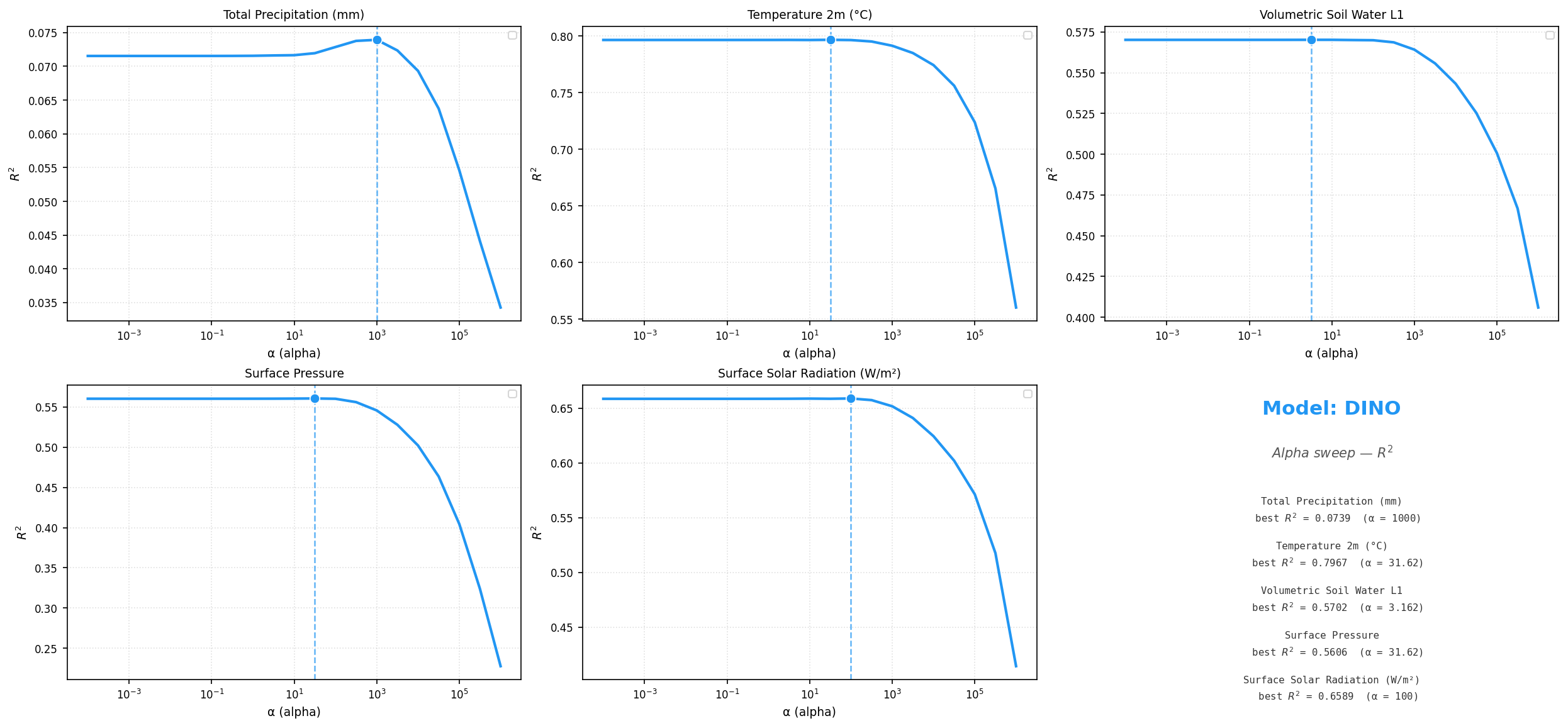}
  \includegraphics[width=0.75\textwidth]{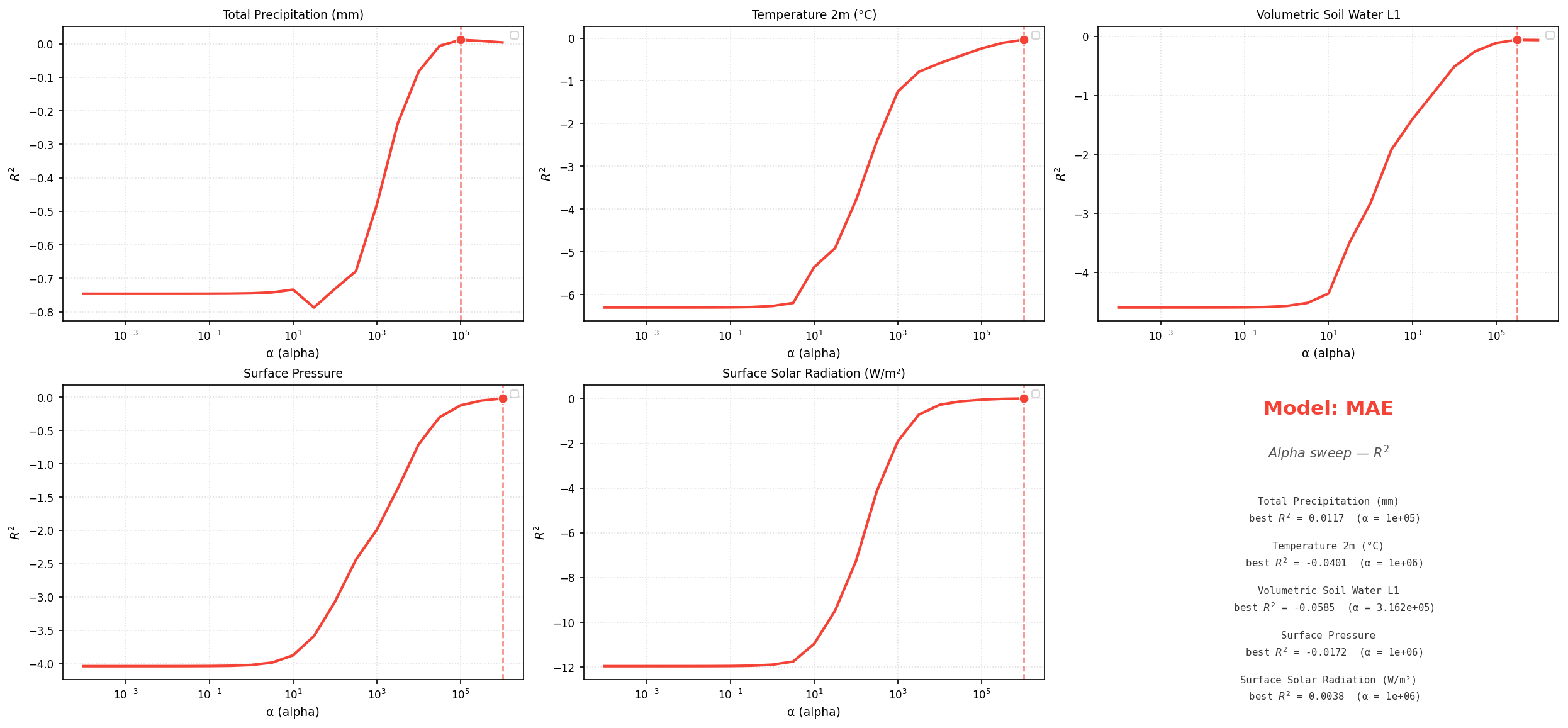}
  \includegraphics[width=0.75\textwidth]{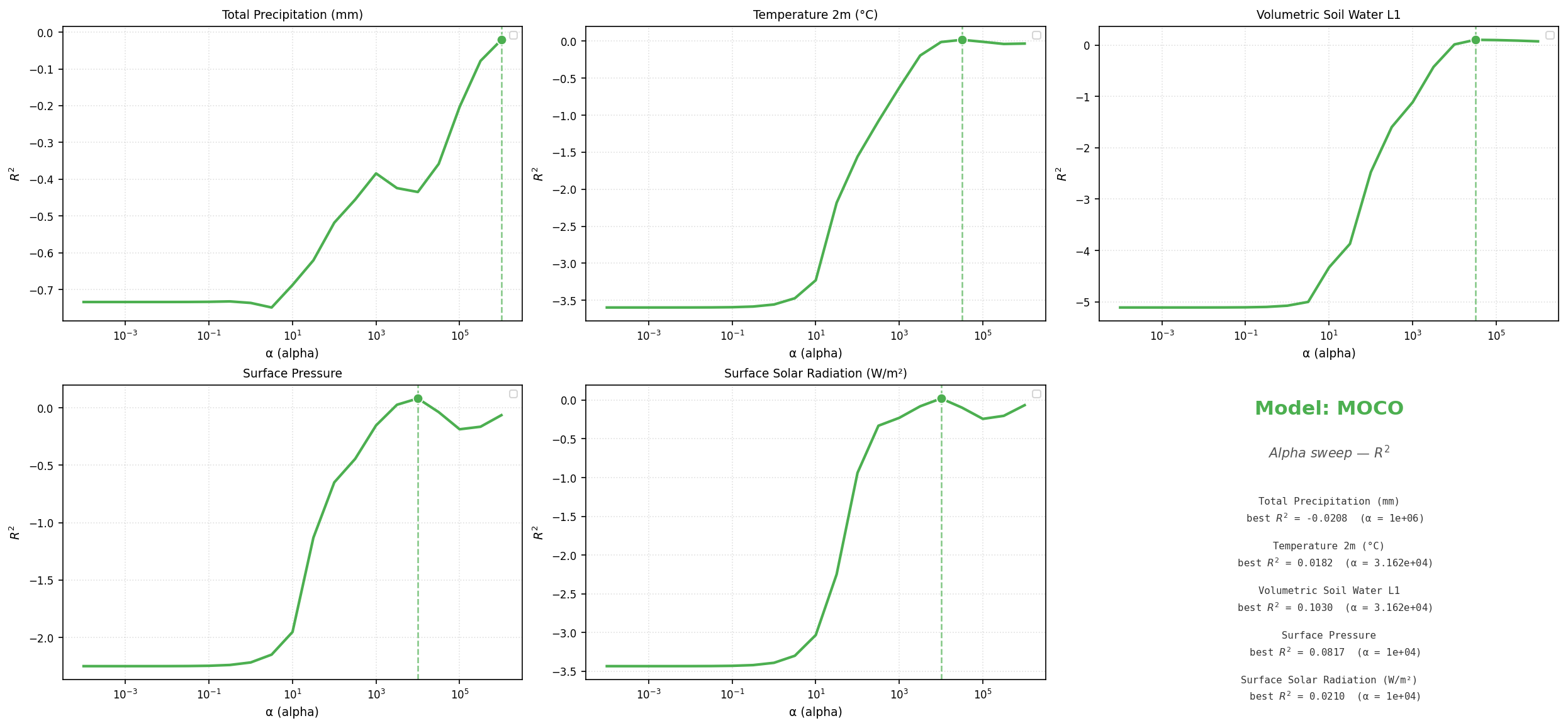}

  \caption{\textbf{Ridge regression hyperparameter search.} We sweep the regularization parameter, $\alpha$, for each SSL objective. Each panel summarizes the highest validation $R^2$ obtained for every ERA5 prediction task together with its corresponding optimal value of $\alpha$.}
    \label{fig:lin_probe}
\end{figure*}

We perform separate hyperparameter searches for the ridge regression and MLP probes. For ridge regression, \cref{fig:lin_probe} shows the sweep over the regularization parameter, $\alpha$, used to identify the best-performing value for each SSL objective. Because the optimal regularization differs slightly across objectives, the selectivity experiments use independently selected values of $\alpha$, while all foundation model comparisons use a fixed regularization parameter ($\alpha=10^4$) to ensure a consistent evaluation protocol. For the MLP probe, \cref{fig:mlp_probe} shows the hyperparameter search over learning rate, hidden dimension, and dropout rate. The selected configuration corresponds to the highest mean $R^2$ performance across the five ERA5 prediction tasks.

\FloatBarrier
\clearpage
\subsection{Out of Distribution Graphs}
\label{app:ood_analysis}
\begin{figure}[ht]
\includegraphics[width=1\columnwidth]{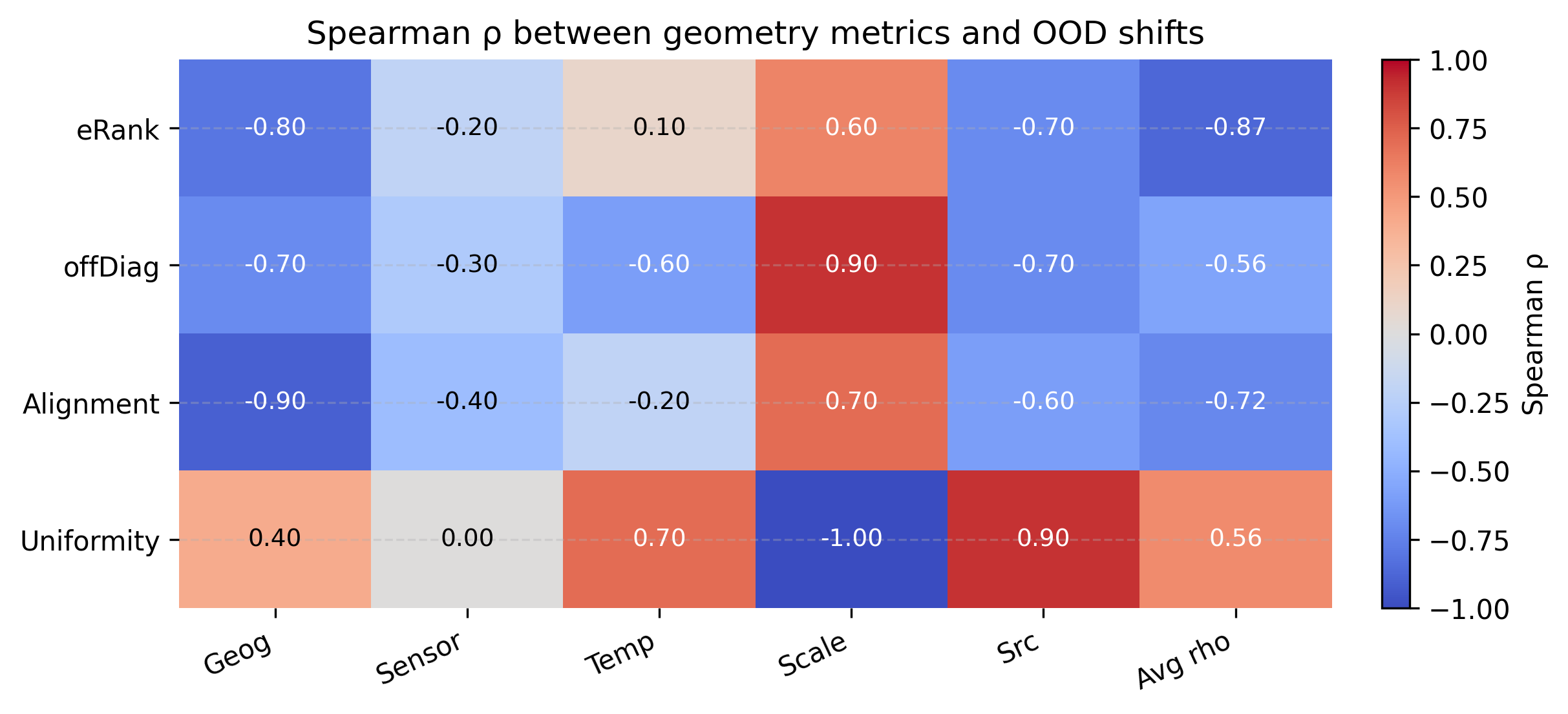}
\centering
\caption{\textbf{Preliminary OOD analysis.} The heatmap summarizes the correlation between the four intrinsic metrics and the five distribution shifts evaluated by EarthShift \cite{doerksen2026earthshift}.}
\label{fig:geometry_ood_heat}
\end{figure}

We perform a preliminary analysis to investigate the relationship between the intrinsic representation metrics and out-of-distribution (OOD) robustness measured by EarthShift \cite{doerksen2026earthshift}. The analysis is conducted using five publicly available foundation models. \Cref{fig:geometry_ood_heat} summarizes the Spearman correlation between the five EarthShift \cite{doerksen2026earthshift} distribution shifts and the four intrinsic representation metrics. While several moderate correlations are observed, the small sample size (n=5) limits the strength of any conclusions.

